\newcommand*{\mytitlelong}{A Latent-Variable Lattice Model}
\newcommand*{\mytitlebroken}{A Latent-Variable Lattice~Model}
\newcommand*{\mytitleshort}{\mytitlelong{}}

\newcommand*{\myemail}{masatran\raisebox{-1pt}{\includegraphics{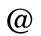}}\!freeshell\,$\cdot$\,org}

\newcommand*{\myuniversity}{Indian Institute of Technology Madras}
\newcommand*{\mydepartment}{Computer Science and Engineering}

\newcommand*{\mykeywords}{Markov random field $\cdot$ Latent variable $\cdot$ Associativity $\cdot$ Inertia}

\mathchardef\mathhyphen="2D
\newcommand*{\norm}[1]{\lVert#1\rVert}

\newcommand*{\citex}[1]{\cite{#1}}

\newcommand{\expectation}[1]{{\mathbb E}\left[ #1 \right]}

\newcommand*{\figurewide}{0.8\columnwidth}

\newcommand*{\tightcaption}{-1em}

\newcommand*{\beginalgo}{\begin{algorithm2e}[!htb]}

\newcommand*{\myalgorithmlearning}{Algorithm~4}

\newcommand*{\algosettings}{ \DontPrintSemicolon \SetKwComment{tcp}{}{} \SetCommentSty{uline} }

    \documentclass[10pt]{article}
    \usepackage[accepted]{_icml2016-arxiv}
    \usepackage{times}
    \icmltitlerunning{\mytitleshort{}}

\usepackage[utf8]{inputenc}
\usepackage{newunicodechar}
\newunicodechar{ā}{\={a}}

\usepackage[algo2e,ruled]{algorithm2e}
\usepackage{amsmath}
\usepackage{amsfonts}
\usepackage{amsopn}
\usepackage{capt-of}
\usepackage{graphicx}
\usepackage{setspace}
\usepackage{xcolor}
\usepackage[normalem]{ulem}
\usepackage{hyphenat}

\usepackage{enumitem}
\setlist{itemsep=0pt}

\DeclareMathOperator{\increment}{increment}
\DeclareMathOperator{\decrement}{decrement}

\DeclareMathOperator{\minindex}{min\,index}

\usepackage{listings}

\usepackage{natbib}

\begin{document}

\title{\mytitlebroken{}}

\twocolumn[
\icmltitle{\mytitlelong{}}
\icmlauthor{Rajasekaran Masatran}{\myemail{}}
\icmladdress{\mydepartment{}, \myuniversity{}}
\icmlkeywords{\mykeywords{}}
\vskip 0.3in
]

\begin{abstract}

Markov random field (MRF) learning is intractable, and its approximation
algorithms are computationally expensive. We target a small subset of MRF that
is used frequently in computer vision. We characterize this subset with three
concepts: Lattice, Homogeneity, and Inertia; and design a non-markov model as
an alternative. Our goal is robust learning from small datasets. Our learning
algorithm uses vector quantization and, at time complexity $O(U \log{U})$ for a
dataset of $U$ pixels, is much faster than that of general-purpose MRF.

\end{abstract}

\section{Introduction}

In computer vision, the MRF is commonly used to model images and videos. MRF
algorithms are intractable in the general case, and approximation algorithms
are generally used. The approximation algorithms themselves are computationally
expensive. For computer vision, we usually do not need the full expressive
power of MRF. In this paper, we restrict ourselves to a small subset of MRF
that is useful for computer vision, and develop fast algorithms that work for
this restricted case. We have three restrictions:

\begin{description}
    \item[Lattice] An image is a two-dimensional lattice.
    \item[Homogeneity] The potentials and conditional probabilities are homogeneous across the lattice.
    \item[Inertia] Nearby nodes are likely to be in the same state.
\end{description}

\subsection{Markov Random Field}

$G = (V, E)$ denotes an undirected graph. $V$ is the set of nodes and $E$ is
the set of undirected edges. $X_{i}$ denotes the variable associated with node
$i$, for $i \in V$; giving a random vector $X = \{ X_{1}, \cdots, X_{p} \}$.
The local markov property for variable $X_{i}$ states that $X_{i}$ given its
set of neighbors $X_{N(i)}$ is conditionally independent of the rest of the
variables. A markov random field over a graph $G$ is the set of distributions
which satisfies all markov properties of $G$.
$P(X) = \frac{1}{Z} \prod_{C \in \mathbb{C}}{\phi_{C}(X_{C})}$ where $C$
is a clique, and $\mathbb{C}$ is the set of cliques. $\phi_{C}(X_{C})$ is the
potential function of clique $C$, and $Z$ is the normalization factor.

\subsection{Lattice}

We restrict our domain to $d$-dimensional lattices. Without loss of generality,
we assume that the lattice has the same length, $T$ nodes, in each dimension.
Each node $t$ is an element of $\{1 \cdots T\}^{d}$. We denote the lattice by
$T^{d}$. These $T^{d}$ nodes are latent variables and are hidden. Each of these
has an associated visible variable that is not counted among its neighbors. The
value of each visible variable is dependent solely on its latent variable.
Given the values of all latent variables, the visible variables are independent
of each other. $q_{t}$ is the state of node $t$.

\begin{description}
    \item[$S = \{ S_{1}, S_{2}, \cdots S_{j}, \cdots S_{N} \}$] is the set of states.
    \item[$Q = \{q_{t}\}$] is the state of all latent variables.
    \item[$O = \{o_{1}, o_{2}, \cdots o_{T}\}$] is the set of observation symbols.
    \item[$R(t)$] is the set of $2 \, d$ nodes adjacent to node $t$.
    \item[$U = \sum{T^{d}}$] is the set of all pixels.
\end{description}

\subsection{Homogeneity}

Due to the markov property, $P(q_{t} | Q \setminus q_{t}) = P(q_{t} |
Q(R(t)))$. We consider only the case where $\phi(t, r \in R(t))$, as well as
$P(o_{t} = v_{k} | q_{t} = S_{j})$ are independent of $t$. Therefore, we
represent $\phi(t, r \in R(t))$ by a common matrix $A$, and
$P(o_{t} = v_{k} | q_{t} = S_{j})$ by a common matrix $B$.

\begin{description}
    \item[$N$] $= \lvert S\rvert$; $S = \{ S_{1}, S_{2}, \cdots S_{j}, \cdots S_{N} \}$ are the unique states in the model, and the state at node~$t$ is $q_{t}$.
    \item[$M$] $= \lvert V\rvert$; $V = \{ v_{1}, v_{2}, \cdots v_{k}, \cdots v_{M} \}$ are the unique observation symbols, and the symbol at node~$t$ is $o_{t}$.
    \item[$A$] $= \{a(i, j)\}$; $a(i, j) = \phi(q_{t} = S_{j}, q_{r \in R(t)} = S_{i})$, the state adjacency potential function.
    \item[$B$] $= \{b(j, k)\}$; $b(j, k) = P(o_{t} = v_{k} | q_{t} = S_{j})$, the observation symbol probability distribution in state $j$.
\end{description}

\subsection{Associativity}

In a latent-variable model, if adjacent latent nodes share the same set of
possible states, and are likely to be in the same state, this is called
associativity. This is an intrinsic property of the process underlying many
application domains. The amount of associativity in the lattice varies by
domain. In the one-dimensional case, there is a lot of associativity in speech
recognition, a limited amount of associativity in chunking of natural language,
and almost no associativity in part-of-speech tagging in natural language.

\setlength{\fboxsep}{2pt}

We propose an index for associativity:
\fbox{$\sum_{i}{a_{i, i}} / \sum_{i, j}{a_{i, j}}$}. It is $0$ when adjacent
nodes are uncorrelated, and $1$ when adjacent nodes are always in the same
state. Our model should be applicable when this index is atleast, say, $0.5$.

\subsection{Inertia}

Associativity is restricted to dependency between adjacent latent nodes. In
this paper, we consider a broader condition, where non-adjacent latent nodes
are dependent on each other. This is not markov, but is related to an
$n^{th}$-order markov model. We call this inertia, and restrict our model to
lattices with inertia. Since an inertial model can approximate an associative
model, it can be used for: (1)~Applications with inertia in the lattice, as
well as (2)~Approximation of applications with associativity in the lattice.

\setlength{\fboxsep}{2pt}

We use a sliding hypercube window to propose an index for inertia:
\fbox{$\expectation{\sqrt{\sum_{j}{p_{j}}^{2}}}$}. The expectation is taken
over the hypercube centered at the node, for every node in the lattice. $p_{j}$
is the probability of state $S_{j}$ in the window. It is $1$ when all nodes in
the window are in the same state, and $\frac{1}{\sqrt{N}}$ when all states have
equal probability within the window. Our model should be applicable when this
index is atleast, say, $0.9$.

\subsection{Examples}

In the following figures, each circle is a MRF node. The thick circles are
latent-variable nodes. Figure~1 is an associative MRF, while Figure~2 is a
non-associative MRF. The matrices are pairwise potentials, and the main
diagonal represents adjacent nodes being in the same state. All nodes, latent
and visible, are boolean. Differences between the two MRFs are highlighted in
bold.

\begin{center}
    \label{fig:mrf-associative}
    \vspace{\tightcaption}
    \includegraphics[width=\figurewide]{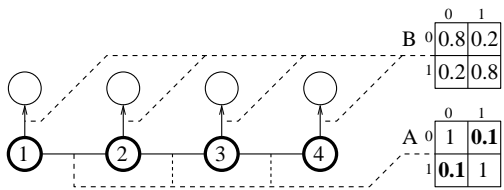}
    \captionof{figure}{Associative MRF}
    \vspace{\tightcaption}
\end{center}

Figure~1 is an associative MRF. It has parameters:
$S$: $\{0, 1\}$, $V$: $\{0, 1\}$, $A$: $\bigl( \begin{smallmatrix} 1 & \mathbf{0.1} \\ \mathbf{0.1} & 1 \end{smallmatrix} \bigr)$,
$B$: $\bigl( \begin{smallmatrix} 0.8 & 0.2 \\ 0.2 & 0.8 \end{smallmatrix} \bigr)$.
The index of associativity is: \\
$\frac{1 + 1}{1 + 0.1 + 0.1 + 1} = 0.91$.

\begin{center}
    \label{fig:mrf-nonassociative}
    \vspace{\tightcaption}
    \includegraphics[width=\figurewide]{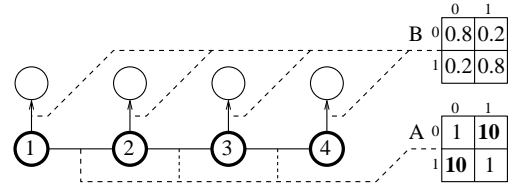}
    \captionof{figure}{Non-Associative MRF}
    \vspace{\tightcaption}
\end{center}

Figure~2 is an non-associative MRF. It has parameters:
$S$: $\{0, 1\}$, $V$: $\{0, 1\}$, $A$: $\bigl( \begin{smallmatrix} 1 & \mathbf{10} \\ \mathbf{10} & 1 \end{smallmatrix} \bigr)$,
$B$: $\bigl( \begin{smallmatrix} 0.8 & 0.2 \\ 0.2 & 0.8 \end{smallmatrix} \bigr)$.
The index of associativity is: \\
$\frac{1 + 1}{1 + 10 + 10 + 1} = 0.09$.

\subsection{Using Stochastic Models}

An assignment of observed symbol to each visible variable is an image. The
image is essentially the visible-variable lattice, as opposed to the
latent-variable lattice. Our training data consists of images, and lacks the
state configuration of the corresponding latent variables.

In classification of images, the input is a set of classes that are specified
by a set of images belonging to each. The objective is to classify test images
into these classes. We build a model for each class by learning a
class-conditional density for its lattice configurations. Given the prior
probabilities and these class-conditional models, bayesian classification is
used to classify test images.

\subsection{Prior Work}

\emph{Graphical models for general networks} were introduced in
\citex{pearl-1988-probabilistic}. \citex{blake-2011-markov} discussed MRF
applications in computer vision.
\emph{In Physics}, the Ising model is widely used. The Potts model, described
in \citex{wu-1982-potts}, is a generalization of the Ising model, and is
closely related to MRF.
\emph{Our main benchmark} is the algorithm for fast inference in associative
pairwise MRF in \citex{boykov-2001-fast}. We solve a slightly different
problem, and our solution should be useful in approximating a solution to
associative MRF as well.

\subsection{Contributions}

\begin{enumerate}
    \item \emph{A latent-variable statistical model for lattice-structured data}: a variant each, for the discrete case (Section 2) and the real case (Section 3)
    \item An index for associativity (Subsection 1.4)
    \item An index for inertia (Subsection 1.5)
\end{enumerate}

\section{Our Model}

In this section, we describe the variant of our model for discrete output
symbols. The one-dimensional case of this variant can be used to model, for
instance, proteins, which are sequences of twenty amino acids.

The signature, discussed in section~3.3, maps the node into the probability
simplex. We divide the probability simplex into partitions, and a partition
corresponds to a state in MRF. The potential between partitions is $A$, and the
observation symbol probability distribution is $B$. Roughly, this is our data
model. Our model is not markov. It has inertia. The parameter set of our model
is $\eta = \langle A, B, w \rangle$.

\subsection{Probability Simplex}

The range of the probability mass for a multinomial distribution with $M$
categories is the is the standard $(M - 1)$ simplex,
$\left\{(p_{1}, p_{2}, \cdots p_{n}) \in \mathbb{R}^{M} \mid \sum_{k = 1}^{n}{p_{k}} = 1, p_{k} \ge 0 \right\}$.
The states of a MRF can be considered points in the probability simplex. The
coordinates of the state are the observation symbol probability distribution
$B$.

\subsection{Sliding Window}

\begin{center}
    \label{fig:sliding-window}
    \includegraphics[width=\figurewide]{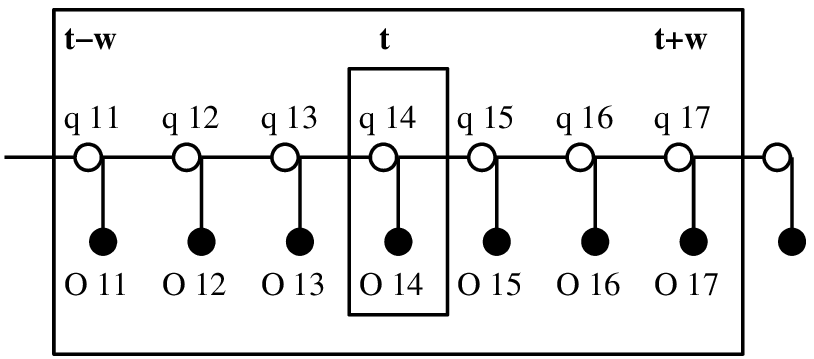}
    \vspace{\tightcaption}
\end{center}

Sliding window methods are popular for sequences
\citex{dietterich-2002-machine} and images. Since our model has inertia, nearby
nodes are probably in the same state. Therefore, sliding window models are
likely to work well. At $t$, the $w$-window is the hypercube from
$(t - [w, w, \cdots])$ to $(t + [w, w, \cdots])$. It might be beneficial to use
different values of $w$ for evaluation, decoding and learning: $w_{e}$ for
evaluation, $w$ for decoding, and $w_{l}$ for learning.

\subsection{Bias-Variance Tradeoff}

The optimal size of the window is related to the typical minimum number of
steps in the same direction before a change of state. Increasing the window
size increases inductive bias, and reduces variance. Reducing the window size
reduces inductive bias, and increases variance.

\subsection{Signature}

Our decoding algorithm works in two steps: (1)~Characterize the sliding window
around the current node with a signature, and (2)~Use this signature to assign
a latent state to the current node. The signature $X$ is that observation
probability vector with maximum likelihood of generating the symbols in the
sliding window. $X = \{x_{t}(k)\}$ where $x_{t}(k)$ is the sample probability
of symbol $v_{k}$ in the $w$-window around node~$t$. This signature is a point
in the probability simplex.

\subsection{Algorithms and their Dependencies}

\begin{center}
    \label{fig:dependency-graph}
    \includegraphics[width=\figurewide]{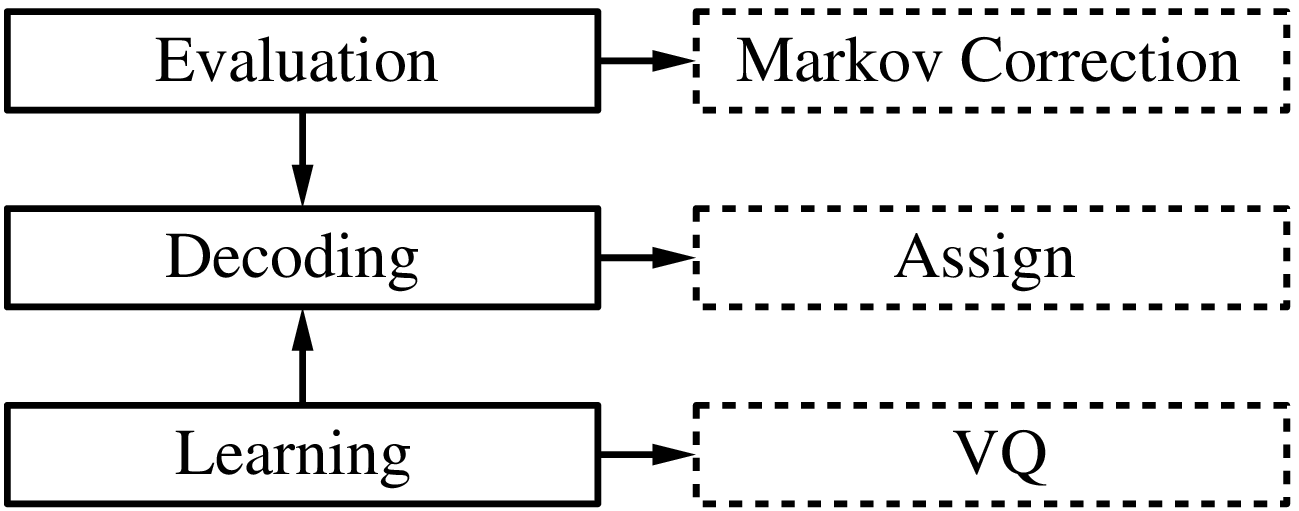}
\end{center}

\subsection{Evaluation}

\emph{Given the model $\eta = \langle A, B, w \rangle$ and the image $O$,
compute the probability $P(O|\eta)$ that the image was produced by the model.}

\beginalgo
\caption{Evaluation ($O(M N T^{d})$)}
    \algosettings
    \KwIn{$A, B, w_{e}, O$}
    \KwOut{$p$}

    $(X, Q) \gets decoding(B, w_{e}, O)$
    \tcp*[f]{$O(M N T^{d})$}\;

    $p  \gets 1$\;

    \For{$t \in T^{d}$}{
        $k \gets \sum_{r \in R(t)}{a(q_{t}, q_{r})}$\;
        $p  \gets p  \; b(q_{t}, o_{t}) \; \sqrt{\prod_{r \in R(t)}{\alpha(X, t) \; \frac{a(q_{t}, q_{r})}{k} }}\mspace{-20mu}$\; 
        \tcp*[f]{Represent in log domain to prevent underflow.}\;

    }
\end{algorithm2e}

Algorithm~1: Since we have inertia, the results of the decoding algorithm are
reliable, and we do not need to consider all possible lattice configurations.
We decode the lattice configuration from the image, and use that to compute the
likelihood.

\subsection{Markov Correction}

$0 < \alpha(X, t) \le 1$ corrects for assuming the markov property in the
evaluation step though it is not present in this model. Experimental evaluation
is required to determine typical values.

\subsection{Decoding}

\emph{Find the lattice configuration $Q = \{q_{t}\}_{t \in T}$, given the model
$\eta = \langle A, B, w \rangle$ and the image $O$.}

\beginalgo
\caption{Decoding ($O(M N T^{d})$)}
    \algosettings
    \KwIn{$B, w, O$}
    \KwOut{$X, Q$}

    $\operatorname{Initialize}_{v_{k} \in V}{symcount(v_{k})}$\;

    \For(\tcp*[f]{$O(M N T^{d})$}){$t \in T^{d}$}{
        $x_{t} \gets \frac{symcount}{M}$\;
        $q_{t} \gets assign(B, x_{t})$
        \tcp*[f]{$O(M N)$}\;
        $\increment{symcount(newwindow - oldwindow)}$\;
        $\decrement{symcount(oldwindow - newwindow)}$\;
    }
\end{algorithm2e}

Algorithm~2: To decode the lattice, we initialize the sliding window and
compute the symbol counts (\emph{symcount}) in it. We slide the sliding window
through the lattice, and in every step, update the symbol counts, and send the
signature of the sliding window to \emph{assign} to get the node state.

\subsection{Assign}

\emph{The assign subroutine finds the node state $q_{t} \in S$, given the model
$\eta = \langle A, B, w \rangle$ and the node signature $x_{t}$.}

\beginalgo
    \caption{Assign ($O(M N)$)}
    \algosettings
    \KwIn{$B, x_{t}$}
    \KwOut{$q_{t}$}

    \For(\tcp*[f]{$O(M N)$}){$j \gets 1$ \KwTo $N$}{
        $dist(j) \gets \norm{x - B(j)}$\;
    }

    $q_{t} \gets \minindex_{j}(dist_{j})$\;

\end{algorithm2e}

Algorithm~3: Since the model has inertia, we reduce computational complexity of
assigning state by assuming that neighboring nodes probably have the same state
as the current node.

The node signature $x_{t}$ is a probability distribution. The $assign$
subroutine assigns $t$ to that state $j$ for which $B(j)$ is the closest to
$x_{t}$, as per $L_{2}$ norm. This creates a partitioning of the probability
simplex into partitions, each of which is associated with a state.

\subsection{Learning}

\emph{Optimize the model parameters $\eta = \langle A, B, w \rangle$ to best
describe the image, i.e., maximize $P(O|\eta)$.}

\beginalgo
    \caption{Learning ($O(M U \log{U})$)}
    \algosettings
    \KwIn{$O, w_{l}$}
    \KwOut{$A, B$}

    $\operatorname{Initialize}_{v_{k} \in V}{symcount(v_{k})}$\;

    \For(\tcp*[f]{$O(M U)$}){$t \in U)$}{
        $x_{t} \gets \frac{symcount}{M}$
        \tcp*[f]{$O(M)$}\;
        $\increment{symcount(newwindow - oldwindow)}$\;
        $\decrement{symcount(oldwindow - newwindow)}$\;
    }

    $(B, Q) \gets VQ(X)$
    \tcp*[f]{$O(M U \log{U})$}\;

    \tcp*[f]{$B$ is assigned the VQ codebook.}\;

    \For(\tcp*[f]{$O(U)$}){$t \in U$}{
        $\increment{statecount(q_{t})}$\;
        \For{$r \in R(t)$}{
            $\increment{A'(q_{t}, r)}$\;
        }
    }

    \For{$j \gets 1$ \KwTo $N$}{
        $A(j) \gets \frac{A'(j)}{2 \; d \; statecount(j)}$\;
    }
\end{algorithm2e}

\myalgorithmlearning{}: Since we have inertia, our decoding algorithm is
reasonably reliable. We decode the image, and use the lattice configuration to
learn the parameters of the model.

We compute the signature at all nodes using symbol counts (\emph{symcount}).
Vector quantization on the signatures partitions the probability simplex into
states. The coordinates of the centroids are the observation probability
matrix, and the sample transition probabilities in the lattice of signatures
are the state potential matrix.

\subsection{Vector Quantization}

Our requirements for clustering are: (1)~Roughly spherical partitions, and
(2)~A representative point for each partition. We solve our clustering problem
as vector quantization (VQ) since the requirements are similar. VQ is a
classical technique from signal processing that models probability density
functions by the distribution of prototype vectors. There are multiple VQ
algorithms with different tradeoffs. Fast pairwise nearest neighbor (Fast PNN)
from \citex{equitz-1989-new} is a VQ algorithm that trades modeling accuracy
for reduced computational complexity. We use Fast PNN as it has $O(n \log{n})$
computational complexity, and we do not need high modeling accuracy in VQ.

\subsection{Time Complexity}

\begin{center}
\begin{tabular}{l|l}
               & Our Model          \\ 
    \hline
    Evaluation & $O(M N T^{d})$     \\ 
    Decoding   & $O(M N T^{d})$     \\ 
    Learning   & $O(M U \log{U})$   \\ 
\end{tabular}

\end{center}

\section{Extension to Real Symbols}

In this section, we describe the variant of our model for real-valued, as
opposed to discrete, output symbols. The two-dimensional case of this variant
can model digital images, in grayscale pixel value, or in extracted features.
Previously, the state variable took values in the probability simplex embedded
in $\mathbb{R}^{M}$. Now, it consists of $M$ real variables representing the
mean vector, i.e., it is a $M$-dimen\-sional vector and has range
$\mathbb{R}^{M}$. $o_{t} = \{o_{t, 1}, o_{t, 2}, \cdots o_{t, M}\}$. The
parameter set of this variant is $\eta = \langle A, \mu, \Sigma, w \rangle$.

When each point is assigned to the closest state, the state space gets divided
into partitions, each of which corresponds to a state in MRF. Matrix $A$
remains the probability of transitioning between partitions, and $w$ remains
the size of the sliding window. We modify the other parameters of our model as
follows:

\begin{description}
    \item[$M$] $= \lvert V\rvert$ was the number of unique observation symbols. Instead, now the observation symbol $o_{t}$ is in $\mathbb{R}^{M}$.
    \item[$B$] was the observation symbol probability distribution. It is replaced by $\mu$ and $\Sigma$.
    \item[$\mu$] $= \mu(j, k)$; where $\mu(j)$ is the sample mean of state $S_{j}$.
    \item[$\Sigma$] $= \Sigma(j, i, k)$; where $\Sigma(j)$ is the sample covariance matrix of state $S_{j}$.
\end{description}

\subsection{Prior Work}

Real-symbol HMM is discussed in subsection IV-A of
\citex{rabiner-1989-tutorial}. Switching linear dynamical systems (LDS) are an
extension of HMM in which each state is associated with a linear dynamical
process. \citex{fox-2011-bayesian} describes inference in switching LDS and
switching vector autoregressive (VAR) processes.

\subsection{Signature}

The signature is the estimated mean of the distribution that generated the
symbols in the sliding window. In the $w$-window around node~$t$, the signature
is the sample mean $x_{t}$ of the symbols in the sliding window.

\subsection{Evaluation}

\emph{Given the model $\eta = \langle A, \mu, \Sigma, w \rangle$ and the image
$O$, compute the probability $P(O|\eta)$ that the image was produced by the
model.}

\beginalgo
\caption{Evaluation ($O(M N T^{d})$)}
    \algosettings
    \KwIn{$A, \mu, \Sigma, w_{e}, O$}
    \KwOut{$p$}

    $(X, Q) \gets decoding(\mu, \Sigma, w_{e}, O)$
    \tcp*[f]{$O(M N T^{d})$}\;

    $p  \gets 1$\;

\For(\tcp*[f]{$O(M^{2} T^{d})$}){$t \in T^{d}$}{
        $k \gets \sum_{r \in R(t)}{a(q_{t}, q_{r})}$\;
        $p  \gets p  \; f(o_{t} | \mu(q_{t}), \Sigma(q_{t})) \; \sqrt{\prod_{r \in R(t)}{\alpha(X, t) \; \frac{a(q_{t}, q_{r})}{k} }}\mspace{-20mu}$\; 
        \tcp*[f]{$f(o | \mu, \Sigma)$: Normal distribution.}\;
        \tcp*[f]{Represent in log domain to prevent underflow.}\;
    }
\end{algorithm2e}

Algorithm~5: Since the clusters generated by VQ all have the same size, there
is no prior probability term in the probability update.

\subsection{Decoding}

\emph{Find the lattice configuration $Q = \{q_{1}, q_{2}, \cdots q_{T}\}$,
given the model $\eta = \langle A, \mu, \Sigma, w \rangle$ and the image $O$.}

\beginalgo
\caption{Decoding ($O(M N T^{d})$)}
    \algosettings
    \KwIn{$\mu, \Sigma, w, O$}
    \KwOut{$X, Q$}

    $\operatorname{Initialize}{sum}$\;

    \For(\tcp*[f]{$O(M N T^{d})$}){$t \in T^{d}$}{
        $x_{t} \gets \frac{sum}{M}$\;
        $q_{t} \gets assign(\mu, \Sigma, x_{t})$
        \tcp*[f]{$O(M N)$}\;
        $\increment{sum(newwindow - oldwindow)}$\;
        $\decrement{sum(oldwindow - newwindow)}$\;
    }
\end{algorithm2e}

Algorithm~6: We initialize the sliding window and compute the sum of the output
instances in it. We slide the sliding window to the end, and in every step, we
update the sum and send the signature of the sliding window to \emph{assign} to
get the state lattice.

\subsection{Assign}

\emph{The assign subroutine finds the state $q_{t} \in S$, given the model
$\eta = \langle A, \mu, \Sigma, w \rangle$, and a node signature $x_{t}$.}

\beginalgo
    \caption{Assign ($O(M N)$)}
    \algosettings
    \KwIn{$\mu, x_{t}$}
    \KwOut{$q_{t}$}

    \For(\tcp*[f]{$O(M N)$}){$j \gets 1$ \KwTo $N$}{
        $dist(j) \gets \norm{x_{t} - \mu(j)}$\;
    }

    $q_{t} \gets \minindex_{j}(dist_{j})$\;

\end{algorithm2e}

Algorithm~7: The $assign$ subroutine assigns $t$ to that state $j$ for which
$\mu(j)$ is the closest to $x_{t}$.

\subsection{Learning}

\emph{Optimize the model parameters $\eta = \langle A, \mu, \Sigma, w \rangle$
to best describe how an image comes about, i.e., maximize $P(O|\eta)$.} (See
Algorithm~8)

\beginalgo
\caption{Learning ($O(M U \log{U})$)}
    \algosettings
    \KwIn{$O, w_{l}$}
    \KwOut{$A, \mu, \Sigma$}

    $\operatorname{Initialize}{sum}$\;

    \For(\tcp*[f]{$O(M U)$}){$t \in U$}{
        $x_{t} \gets \frac{sum}{M}$
        \tcp*[f]{$O(M)$}\;
        $\increment{sum(newwindow - oldwindow)}$\;
        $\decrement{sum(oldwindow - newwindow)}$\;
    }

    $(\mu, Q) \gets VQ(X)$
    \tcp*[f]{$O(M U \log{U})$}\;

    \tcp*[f]{$\mu$ is assigned the VQ codebook.}\;

    \For(\tcp*[f]{$O(M^{2} U)$}){$t \in U$}{
        $\increment{statecount(q_{t})}$\;
        \For{$r \in R(t)$}{
            $\increment{A'(q_{t}, r)}$\;
        }
        $y \gets (o_{t} - \mu(q_{t}))$\;
        $\Sigma'(q_{t}) \gets \Sigma'(q_{t}) + y y'$
        \tcp*[f]{$O(M^{2})$}\;
    }

    \For{$j \gets 1$ \KwTo $N$}{
        $A(j) \gets \frac{A'(j)}{2 \; d \; statecount(j)}$\;
        $\Sigma(j) \gets \frac{\Sigma'(j)}{statecount(j)}$\;
    }
\end{algorithm2e}

\subsection{Time Complexity}

The time complexities are the same as the original model. In case $N$ and $M$
are not considered constant, these caveats apply:
(1)~In Algorithm~5, $O(M^{2} T^{d})$ is assumed to be less than $O(M N T^{d})$, since $O(M)$ is usually less than $O(N)$, and
(2)~In Algorithm~8, $O(M^{2} U)$ is assumed to be less than $O(M U \log{U})$, since $O(M)$ is usually less than $O(\log{U})$.

\section{Conclusion}

We have designed an alternative model for a subclass of MRF that is used
frequently in computer vision. We have two variants: one for lattices of
discrete symbols, and one for lattices of vectors. Our learning algorithm, at
time complexity $O(U \log{U})$, is significantly faster than algorithms
of general-purpose MRF. We are currently evaluating our model on the PASCAL VOC
Challenge 2006 \citex{pascal-voc-2006}.

\subsection{Advantages}

\begin{enumerate}
    \item Variance component of error is low.
    \item Uses small datasets efficiently.
    \item Computational complexity is low.
    \item Easily extends to a quadtree approach.
\end{enumerate}

\subsection{Disadvantages}

\begin{enumerate}
    \item Bias component of error is high.
    \item Does not use large datasets efficiently.
    \item Modeling accuracy is low.
    \item $N$, $w$, and $\beta$ need to be chosen appropriately.
\end{enumerate}

\subsection{Future Work}

\begin{enumerate}
    \item A quadtree approach can provide progressive results.
    \item The model can be repeated multiple times, each with a different window size, and combined into a factorial model.
    \item A theoretical framework can be developed for guarantees on modeling accuracy.
    \item Estimating a belief-state instead of a vanilla state might give better accuracy.
    \item $N$ can be learned from the data.
    \item $w$ can be learned from the data.
\end{enumerate}

\vspace{0.5em}
\noindent
\textbf{Acknowledgments:}
Gow\-tha\-man Aru\-mugam. 

    \bibliographystyle{_icml2016-arxiv}

\begin{thebibliography}{9}
\providecommand{\natexlab}[1]{#1}
\providecommand{\url}[1]{\texttt{#1}}
\expandafter\ifx\csname urlstyle\endcsname\relax
  \providecommand{\doi}[1]{doi: #1}\else
  \providecommand{\doi}{doi: \begingroup \urlstyle{rm}\Url}\fi

\bibitem[Blake et~al.(2011)Blake, Kohli, and Rother]{blake-2011-markov}
Blake, Andrew, Kohli, Pushmeet, and Rother, Carsten.
\newblock \emph{{Markov Random Fields for Vision and Image Processing}}.
\newblock MIT Press, 2011.
\newblock ISBN 0262015773, 9780262015776.

\bibitem[Boykov et~al.(2001)Boykov, Veksler, and Zabih]{boykov-2001-fast}
Boykov, Yuri, Veksler, Olga, and Zabih, Ramin.
\newblock {Fast Approximate Energy Minimization via Graph Cuts}.
\newblock \emph{IEEE Transactions on Pattern Analysis and Machine
  Intelligence}, 23\penalty0 (11):\penalty0 1222--1239, 2001.

\bibitem[Dietterich(2002)]{dietterich-2002-machine}
Dietterich, Thomas~G.
\newblock {Machine Learning for Sequential Data: A Review}.
\newblock In \emph{Structural, Syntactic, and Statistical Pattern Recognition},
  volume 2396 of \emph{LNCS}, pp.\  15--30, 2002.

\bibitem[Equitz(1989)]{equitz-1989-new}
Equitz, William~H.
\newblock {A New Vector Quantization Clustering Algorithm}.
\newblock \emph{IEEE Transactions on Acoustics, Speech and Signal Processing},
  37\penalty0 (10):\penalty0 1568--1575, 1989.

\bibitem[Everingham et~al.(2006)Everingham, Zisserman, Williams, and van
  Gool]{pascal-voc-2006}
Everingham, Mark, Zisserman, Andrew, Williams, Chris K.~I., and van Gool, Luc.
\newblock {The PASCAL Visual Object Classes Challenge 2006 Results}, 2006.

\bibitem[Fox et~al.(2011)Fox, Sudderth, Jordan, and Willsky]{fox-2011-bayesian}
Fox, Emily~B., Sudderth, Erik~B., Jordan, Michael~I., and Willsky, Alan~S.
\newblock {Bayesian Nonparametric Inference of Switching Dynamic Linear
  Models}.
\newblock \emph{IEEE Transactions on Signal Processing}, 59\penalty0
  (4):\penalty0 1569--1585, 2011.

\bibitem[Pearl(1988)]{pearl-1988-probabilistic}
Pearl, Judea.
\newblock \emph{{Probabilistic Reasoning in Intelligent Systems: Networks of
  Plausible Inference}}.
\newblock Morgan Kaufmann Publishers, 1988.
\newblock ISBN 0-934613-73-7.

\bibitem[Rabiner(1989)]{rabiner-1989-tutorial}
Rabiner, Lawrence~R.
\newblock {A Tutorial on Hidden Markov Models and Selected Applications in
  Speech Recognition}.
\newblock \emph{Proceedings of the IEEE}, 77\penalty0 (2):\penalty0 257--286,
  1989.

\bibitem[Wu(1982)]{wu-1982-potts}
Wu, Fa-Yueh.
\newblock {The Potts Model}.
\newblock \emph{Reviews of Modern Physics}, 54\penalty0 (1):\penalty0 235--268,
  1982.

\end{thebibliography}

\end{document}